\documentclass{vgtc}                          

\graphicspath{{figures/}{pictures/}{images/}{./}} 
\usepackage{makecell}
\usepackage{times}                     

\usepackage{tabu}                      
\usepackage{booktabs}                  
\usepackage{lipsum}                    
\usepackage{mwe}                       

\usepackage{tabularx}
\usepackage{array}
\usepackage{graphicx}

\usepackage{mathptmx}                  
\usepackage{balance}

\newcommand{\revise}[1]{{\color{black} #1}}

\newcommand{\inlineicon}[2][1em]{%
  \raisebox{-0.15em}{\includegraphics[height=#1]{#2}}%
}

\onlineid{0}

\vgtccategory{Research}

\vgtcinsertpkg

\title{ChartMark: A Structured Grammar for Chart Annotation}

\author{
Yiyu Chen
\thanks{e-mail: ychen081@connect.hkust-gz.edu.cn}
\\ %
        \scriptsize {HKUST(GZ)} %
\and Yifan Wu\\ %
        \scriptsize {HKUST(GZ)} %
\and Shuyu Shen\\ %
        \scriptsize {HKUST(GZ)} %
\and Yupeng Xie\\ %
        \scriptsize {HKUST(GZ)} %
\and Leixian Shen\\ %
        \scriptsize {HKUST} %
\and Hui Xiong\\ %
        \scriptsize {HKUST(GZ)} %
\and Yuyu Luo\thanks{e-mail: yuyuluo@hkust-gz.edu.cn (corresponding author)}\\ %
        \scriptsize {HKUST(GZ)} %
}

\abstract{
Chart annotations enhance visualization accessibility but suffer from fragmented, non-standardized representations that limit cross-platform reuse. We propose ChartMark, a structured grammar that separates annotation semantics from visualization implementations. ChartMark features a hierarchical framework mapping onto annotation dimensions (e.g., task, chart context), supporting both abstract intents and precise visual details. Our toolkit demonstrates converting ChartMark specifications into Vega-Lite visualizations, highlighting its flexibility, expressiveness, and practical applicability.
} 


\teaser{
  \centering
  \includegraphics[width=\linewidth, alt={A view of a city with buildings peeking out of the clouds.}]{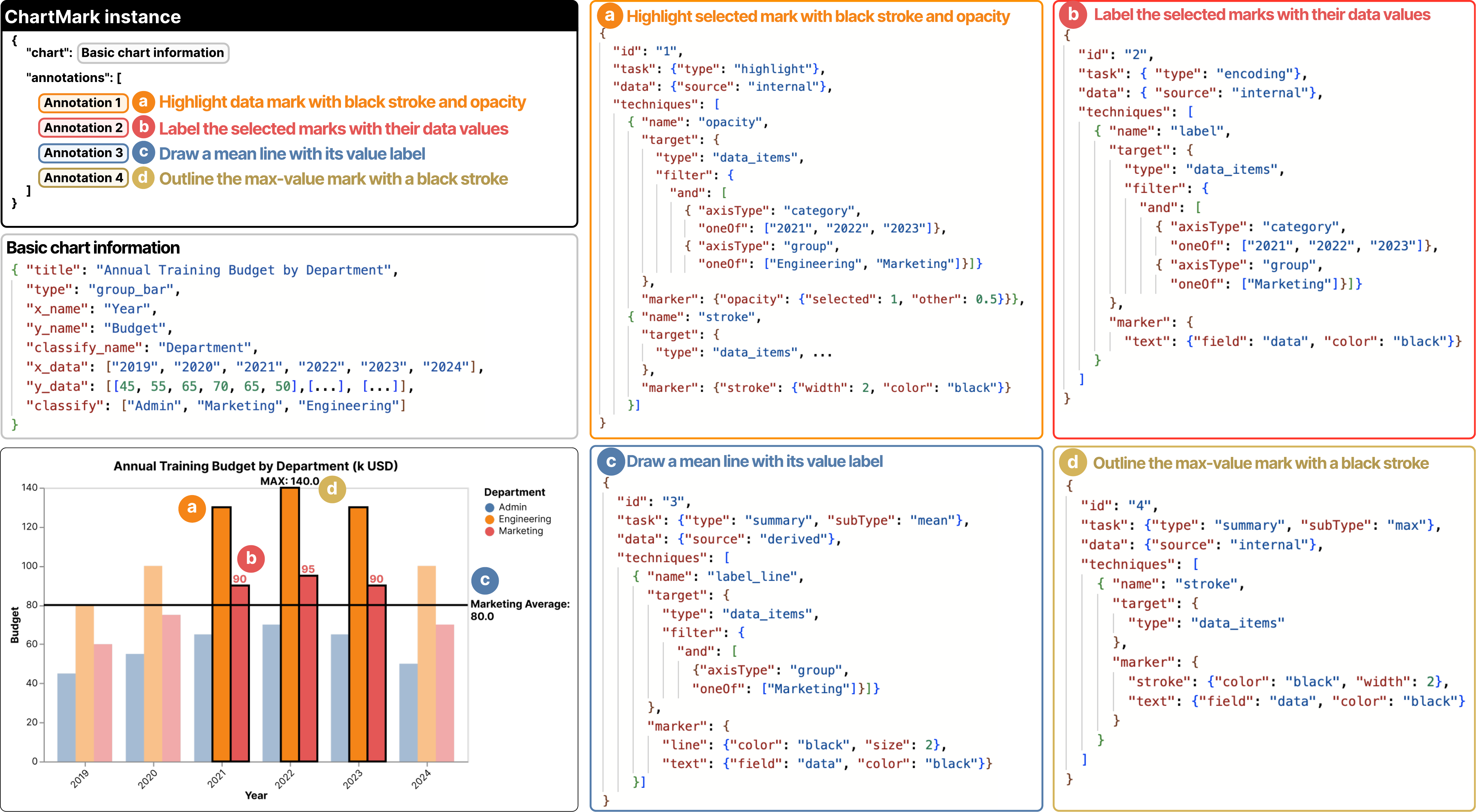}
  \vspace{-1.5em}
  \caption{%
An annotated chart and its corresponding ChartMark grammar representation. Left-top: The overall ChartMark instance showing the annotations list structure. Left-center: Basic chart information defining the group bar chart properties. Left-bottom: The rendered chart displaying annual training budget by department with four annotations: highlighting data marks with stroke and opacity (a), labeling marks with data values (b), drawing a mean line with value label (c), and outlining the max-value mark (d). Center and Right: The detailed JSON specification for each annotation (a-d). 
\revise{For more examples and details, see this
\href{https://chartmark.github.io/}{link}.
}
  }
  \label{fig:teaser}
  \vspace{-0.8 em}
}

\newcommand{\eat}[1]{}
\usepackage[utf8]{inputenc} 
\usepackage[T1]{fontenc}    
\usepackage{hyperref}       
\usepackage{url}            
\usepackage{booktabs}       
\usepackage{amsfonts}       
\usepackage{nicefrac}       
\usepackage{microtype}      
\usepackage{xcolor}         

\usepackage{booktabs}
\usepackage[color,matrix,arrow,all]{xy}
\usepackage{pifont}
\usepackage{url}
\usepackage{listings}
\usepackage[flushleft]{threeparttable}
\usepackage{makecell}
\usepackage{xparse}
\usepackage{wrapfig}
\usepackage{tcolorbox}

\usepackage{epsfig}
\usepackage{multirow}

\usepackage{bbm}

\definecolor{shadecolor}{RGB}{220,220,220}

\definecolor{inputcolor}{RGB}{255,139,35}
\definecolor{outputcolor}{RGB}{120,212,252}
\definecolor{embedcolor}{RGB}{254,127,156}
\definecolor{maskcolor}{RGB}{122,128,255}
\definecolor{ecolor}{RGB}{58,149,54}

\definecolor{highcolor}{RGB}{255,153,153}
\definecolor{midcolor}{RGB}{255,204,204}
\definecolor{lowcolor}{RGB}{204,229,255}

\usepackage{tikz}
\usetikzlibrary{shapes,snakes}
\usetikzlibrary{calc}

\usepackage{algorithm} 
\usepackage{algorithmic}  
\usepackage[algo2e]{algorithm2e} 

\usepackage[export]{adjustbox}

\definecolor{green}{RGB}{0,128,0}

\definecolor{yellow}{RGB}{255,200,18}

\newcommand{\stab}{\vspace{1.2ex}\noindent}

\newcommand{\bi}{\begin{itemize}}
\newcommand{\ei}{\end{itemize}}

\newcommand{\be}{\begin{enumerate}}
\newcommand{\ee}{\end{enumerate}}
\newcommand{\beqn}{\begin{eqnarray*}}
\newcommand{\eeqn}{\end{eqnarray*}}

\newcommand{\stitle}[1]{\stab\noindent{\bf #1}}

\newcommand{\eg}{\textit{e.g.,}\xspace}

\usepackage{tcolorbox}
\usepackage{colortbl}
\usepackage{soul}
\definecolor{c1}{cmyk}{0,0.6175,0.8848,0.1490}
\definecolor{c2}{cmyk}{0.1127,0.6690,0,0.4431}
\definecolor{c3}{cmyk}{0.3081,0,0.7209,0.3255}
\definecolor{c4}{cmyk}{0.6765,0.2017,0,0.0667}
\definecolor{c5}{cmyk}{0,0.8765,0.7099,0.3647}

\newtcbox{\hlprimarytab}{on line, rounded corners, box align=base, colback=c3!10,colframe=white,size=fbox,arc=3pt, before upper=\strut, top=-2pt, bottom=-4pt, left=-2pt, right=-2pt, boxrule=0pt}
\newtcbox{\hlsecondarytab}{on line, box align=base, colback=red!10,colframe=white,size=fbox,arc=3pt, before upper=\strut, top=-2pt, bottom=-4pt, left=-2pt, right=-2pt, boxrule=0pt}
\newtcolorbox[]{finding}[0]{colback=gray!10, colframe=black, width=\columnwidth, boxrule=0.4pt, 
left=0mm, right=0mm, top=0mm, bottom=0mm, before skip=3pt, after skip=3pt, sharp corners}

\makeatletter
    \newcommand\figcaption{\def\@captype{figure}\caption}
    \newcommand\tabcaption{\def\@captype{table}\caption}
\makeatother

\tikzstyle{mybox} = [draw=black, fill=black!5, thick,
   rectangle, rounded corners, inner sep=0pt, inner ysep=6pt]
\tikzstyle{fancytitle} =[fill=black, text=white]

\NewDocumentCommand{\nan}{ mO{} }{\textcolor{blue}{\textsuperscript{\textit{Nan}}\textsf{\textbf{\small[#1]}}}}

\NewDocumentCommand{\yuyu}{ mO{} }{\textcolor{green}{\textsuperscript{\textit{Yuyu}}\textsf{\textbf{\small[#1]}}}}

\NewDocumentCommand{\yifan}{ mO{} }{\textcolor{brown}{\textsuperscript{\textit{Yifan}}\textsf{\textbf{\small[#1]}}}}

\NewDocumentCommand{\yiyu}{ mO{} }{\textcolor{green}{\textsuperscript{\textit{Yiyu}}\textsf{\textbf{\small[#1]}}}}

\NewDocumentCommand{\yupeng}{ mO{} }{\textcolor{purple}
{\textsuperscript{\textit{yupeng}}\textsf{\textbf{\small[#1]}}}}

\usepackage{marginnote}
\let\oldmarginpar\marginpar
\renewcommand\marginpar[1]{\-\oldmarginpar[\raggedleft\footnotesize #1]%
	{\raggedright\footnotesize\color{blue} #1}} 
\usepackage{marginnote}
\let\oldmarginnote\marginnote
\renewcommand\marginnote[1]{\-\oldmarginnote[\raggedleft\footnotesize #1]%
	{\raggedright\footnotesize\color{blue} #1}} %

\begin{document}

\maketitle

\section{Introduction}
Annotations are crucial elements in data visualizations~\cite{wu2025chartcards, wu2024chartinsights,DBLP:conf/icde/LuoQ0018}, directing viewer attention, emphasizing patterns, and providing contextual information~\cite{yang2024askchart, wu2025boosting, bateman2010useful}. Despite their importance, current annotation approaches lack standardization and are often tightly coupled to specific visualization libraries~\cite{Satyanarayan2017,Stolte2002b,Wickham2010}, making them difficult to reuse across platforms.

Previous research has explored annotation spaces along dimensions~\cite{moritz2018formalizing,rahman2024qualitative,shi2025augmenting, chen2022vizbelle} such as data references, annotation tasks, visual forms, and implementation techniques. However, these classifications are typically incomplete or inconsistently defined, capturing only portions of the annotation landscape. Consequently, it remains challenging to create annotation systems that are both semantically rich and versatile enough for diverse applications.

To address these limitations, we present ChartMark, a structured grammar for visual annotations. ChartMark provides a comprehensive and extensible framework that explicitly decouples the semantic representation of annotations from their specific chart implementations. Unlike existing visualization grammars (e.g., Vega-Lite~\cite{Satyanarayan2017}) and annotation tools (e.g., ChartAccent~\cite{ren2017chartaccent} and Charticulator~\cite{ren2018charticulator}), which typically couple annotation specifications tightly to particular visualization environments, ChartMark introduces an independent abstraction layer. This design allows annotations to be consistently reused across multiple visualization platforms without losing their intended semantic meanings.

Our contributions include: (1) a language-agnostic annotation representation framework structured into meaningful semantic layers, (2) the ChartMark grammar for formalizing chart annotations, and (3) a toolkit implementation (\texttt{ChartMark}) demonstrating practical application by transforming ChartMark specifications into Vega-Lite visualizations. These provide foundations for creating more systematic, reusable annotation systems across visualization platforms.

\section{Related Work}

\revise{
\textbf{Annotation in Visualization Grammar.} 
Visualization languages and libraries~\cite{Wickham2010,Bostock2011,Siddiqui2016b,Stolte2002b,xie2024haichart,PyGWalker,taskvis,taskvisDSE,galvis} provide a systematic approach to defining visualizations. Some of them support annotation-like effects to varying extents. 
Languages based on the Grammar of Graphics model \cite{wilkinson2012grammar}, such as D3 
\cite{Bostock2011D3}, Vega \cite{Satyanarayan2014Declarative}, and Vega-Lite \cite{Satyanarayan2017}, do not provide independent rendering modules for annotations and do not distinguish between graphics used for annotations and those used for basic chart elements. Annotation effects can be simulated by layering primitive marks like text and lines on top of existing visual layers. However, this design leads that the annotation structure is tightly coupled with the overall rendering logic. In contrast, languages/libraries like ECharts \cite{Li2018ECharts}, matplotlib \cite{Hunter2007Matplotlib}, and ggplot2 \cite{Wickham2016ggplot2} offer dedicated APIs for annotations. But these APIs provide limited annotation rendering capabilities and lack support for specifying rich, expressive annotations or managing multiple annotations. Despite their differences, annotation mechanisms in both types of systems are heavily dependent on the internal design of their respective grammars. This structural dependency greatly limits the portability and reuse potential of annotations between different visualization systems, forming a technical barrier.
}

\noindent\textbf{Annotation Tools and Platforms.} The aforementioned coupling problem also exists in visualization tools that focus on annotation functions. As a pioneer in interactive chart annotation, ChartAccent~\cite{ren2017chartaccent} provides a rich variety of annotation technologies, but its implementation is still tightly bound to a specific platform. 
In the field of data-driven narrative systems~\cite{amini2016authoring,eccles2008stories}, DataClips~\cite{amini2016authoring}, GeoTime Stories~\cite{eccles2008stories}, and Timeline Storyteller~\cite{brehmer2019timeline} all integrate annotation modes that are critical to their narrative functions, but these annotation mechanisms are only designed for specific platforms and application scenarios, making it difficult to apply in a wider environment. Similarly, although the annotation functions of commercial visualization tools such as Tableau~\cite{tableau} and Adobe Illustrator~\cite{adobe_Illustrator} are powerful, their proprietary implementations for their own products also lead to the non-portability of annotations.
The common limitation of these tools is that they focus on operational functions rather than abstract annotation semantics. Although they can help users create effective annotations, they do not provide standardized methods for representing, exchanging, or reusing these annotations in different visualization environments.

\section{Goals \& Requirements}

A dedicated grammar for chart annotation can address the limitations of existing approaches. This grammar should include the following design considerations:

\revise{
\textbf{DC1: Comprehensive and Extensible 
Semantics.}
The annotation design spaces for specific domains (such as finance \cite{hao2024finflier} or time series charts \cite{hullman2013contextifier}) are limited in expression; while the existing design space solutions \cite{chen2022vizbelle, rahman2024qualitative} for general annotations lack scalability and cannot evolve with new scenarios or new visualization styles. So our annotation representation should cover all possible annotation scenarios in the current design space while remaining flexible enough to accommodate future extensions.
}

\revise{
\textbf{DC2: Language Independence.}
The annotation methods integrated in existing visualization languages \cite{Hunter2007Matplotlib, Satyanarayan2017, Li2018ECharts, Wickham2016ggplot2, Bostock2011D3, Satyanarayan2014Declarative} are not designed specifically for general annotations. Their expression and usage are strictly constrained by the underlying syntax and rendering model. To address this limitation, our annotation grammar should focus on the semantic information of annotations without dependency on any specific implementation methods and rendering languages so that it can achieve broader applicability and cross-platform reuse.}

\revise{
\textbf{DC3: Modular Architecture.}
Chart annotations are semantically additions to the base chart and should be represented as modules independent of the base chart. Each specific annotation should also be considered an independent unit that can be flexibly added or deleted without affecting the original chart structure.
}

\revise{
\textbf{DC4: Multi-level Semantic Representation.}
The expressions of annotations are usually multi-semantic. For example, they include the high-level intention of ``why do it'' (\eg ``Highlight `A' ''), and can also describe the specific operation of ``how to do it'' (\eg ``Set `A' red and fade other marks''). The grammar needs to include these different semantic levels to ensure its expressive completeness.
}

\revise{
\textbf{DC5: Atomic Element Design.}
Basic actions like ``Set Color'', ``Adjust Opacity'' and ``Add Lines'' often reused between different annotation types. To reduce redundancy and enable complex representations, the grammar can view these basic actions or parameter settings as atomic elements that can be reused and combined to create richer annotation expressions.
}

\section{ChartMark Grammar: Design and Formalization}

\begin{table}[t!]
\centering\small
\caption{Dimensions and Components of Chart Annotations}
\begin{tabularx}{\linewidth}{%
  >{\raggedright\arraybackslash}p{1.5cm} 
  |l
  |X}
\hline
\textbf{Dimension} & \textbf{Component} & \textbf{Description} \\ \hline
Data &
  \texttt{Data} &
  The data entities or aggregates that an annotation references. \\ \hline
Task or Goal &
  \texttt{Task} &
  The analytic or communicative goal motivating the annotation (e.g., highlight, compare, explain). \\ \hline
Embellishment or Marker &
  \texttt{Marker} &
  Added or modified visual elements used to convey the annotation. \\ \hline
Operational Target &
  \texttt{Target} &
  The specific marks or regions that the annotation acts upon. \\ \hline
Operation or Technique &
  \texttt{Operation} &
  The concrete methods or tactics (color change, overlay, etc.) employed to realize the annotation. \\ \hline
Chart Context &
  \texttt{Chart} &
  Chart features (typically chart type) that influence annotation display and rendering. \\ \hline
\end{tabularx}
\label{tab:annotation-dimensions}
\end{table}

\begin{figure}[t!]
	\centering	\includegraphics[width=\linewidth]{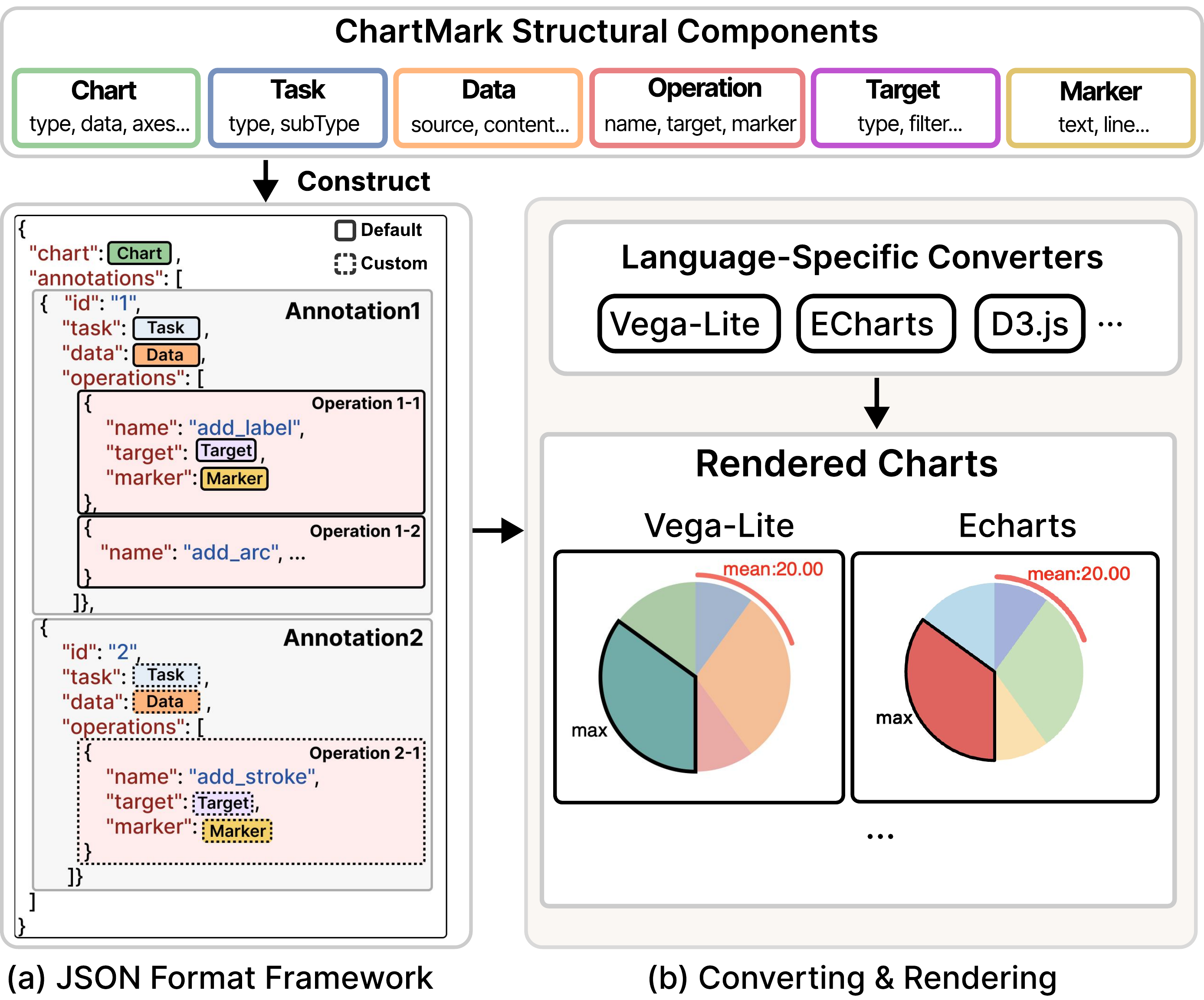}
    \vspace{-1.5em}
    \caption{
    Overview of the ChartMark grammar pipeline. It illustrates how annotations are represented hierarchically through JSON specifications (a), and (b) demonstrates the transformation of these abstract specifications into platform-specific visualizations using converters for Vega-Lite, ECharts, and D3.js.
    } 
    \label{fig:garmmar_overview}
    \vspace{-1.5em}
\end{figure}

Given the design considerations mentioned above, we propose the ChartMark grammar. \cref{fig:garmmar_overview} shows the pipeline of ChartMark from its design to rendering. ChartMark consists of five features: (1) Comprehensive integration of annotation classification methods with extensibility through custom nodes (\textbf{DC1}); (2) separation of semantic information from implementation, allowing conversion to different visualization languages through converters (\textbf{DC2}); 
\revise{ (3) structural separation of annotations from base charts and modular arrangement of annotations (\textbf{DC3});
} 
(4) hierarchical JSON organization to express multiple semantic levels (\textbf{DC4}); and (5) atomic components that serve as building blocks for composite annotations (\textbf{DC5}).

\revise{
The ChartMark specification codes in this paper are simplified for explanation purposes and may omit some details for brevity. Please see the homepage of CharMark (\href{https://chartmark.github.io/}{chartmark.github.io}) for more examples and details.
}

\subsection{Component and Framework}

\revise{
By summarizing the studies on the chart annotation design space~\cite{moritz2018formalizing,rahman2024qualitative,shi2025augmenting,shen2024ask,kong2012graphical,chen2022vizbelle, ren2017chartaccent, hullman2013contextifier, bryan2016temporal},
}
we categorize annotations into multiple dimensions (see~\cref{tab:annotation-dimensions}). In the ChartMark grammar, we design a series of basic components (see~\cref{fig:garmmar_overview}) that directly map to these annotation classification dimensions. Every annotated chart represented in ChartMark must contain and specify these components. This structured approach ensures comprehensive semantic expression while maintaining consistent representation across visualization platforms.

Each classification dimension in~\cref{tab:annotation-dimensions} contains different specific classification values, representing various semantic features of annotations across dimensions, which will be discussed in~\cref{sec:grammar_definition}.

These classification dimensions exhibit both parallel and hierarchical relationships. To effectively represent these relationships, we structure the corresponding components in a hierarchical JSON format (see \cref{fig:garmmar_overview}-a). This JSON structure naturally expresses the logical connections between classification dimensions through its parallel and nested organization:

(1) The outer layer contains the Chart and Annotation list, separating the base chart from the annotation representation, making annotations modular and pluggable.

(2) The middle layer structure contains Task, Data, and Operation lists, representing the data and operation sets used by an annotation to achieve a specific task.

(3) The inner layer structure defines the specific operation Targets and added Marker for each Operation, detailing the specifics of operation execution.

\subsection{Annotation Grammar Definition}
\label{sec:grammar_definition}

\begin{figure}[t!]
	\centering
    \includegraphics[width=\linewidth]{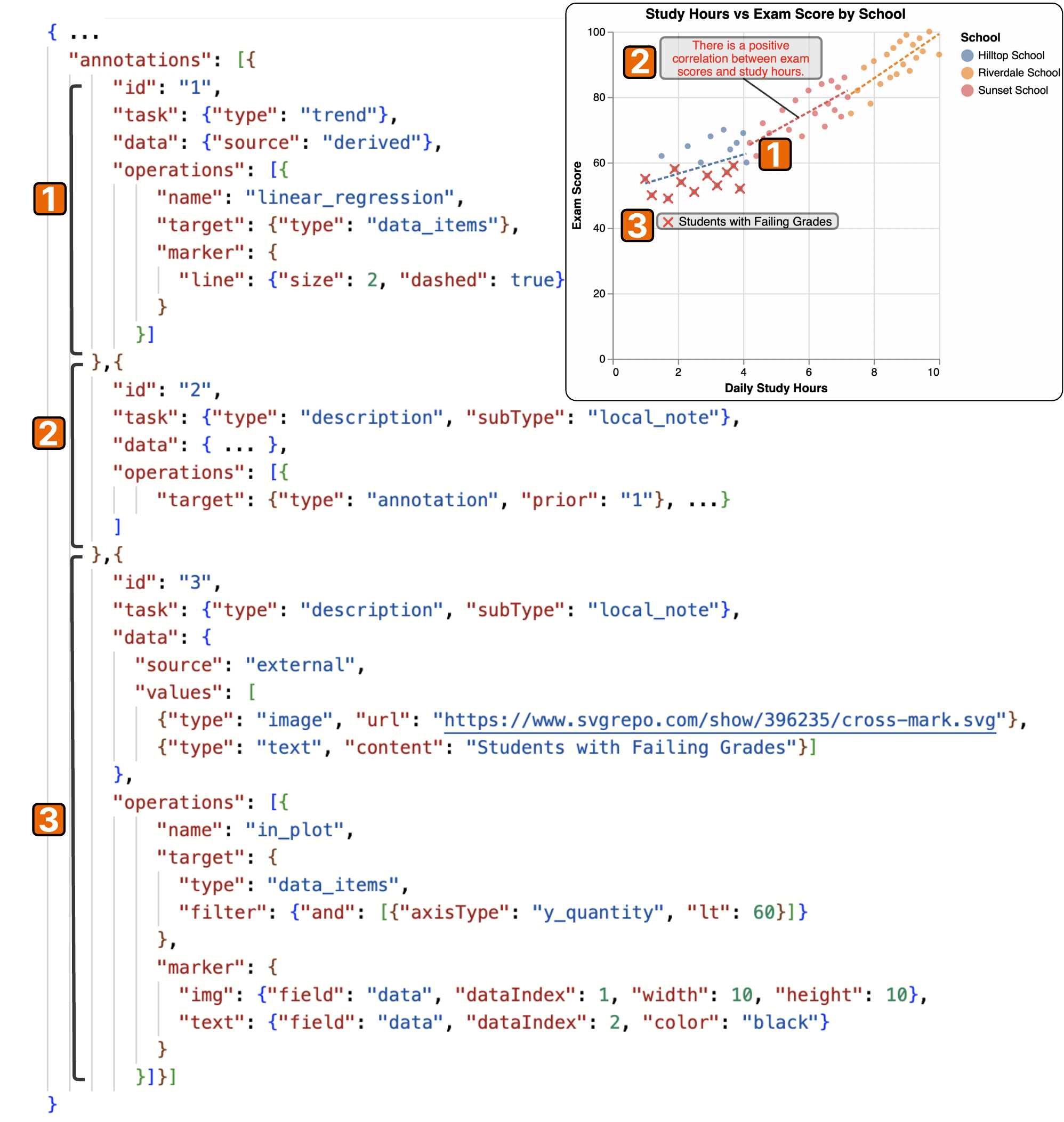}
    \vspace{-1.5em}
    \caption{ 
ChartMark specification for a scatter plot with three annotations: (\inlineicon{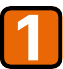}) dashed trend lines using linear regression across school groups, (\inlineicon{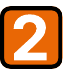}) an explanatory text note attached to the trend line highlighting the positive correlation, and (\inlineicon{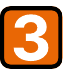}) cross-mark symbols on failing scores (<60) with a legend explaining their meaning.
    } 
    \label{fig:Group_Scatter_Chart_Example}
\end{figure}

\begin{figure}[t!]
	\centering	\includegraphics[width=\linewidth]{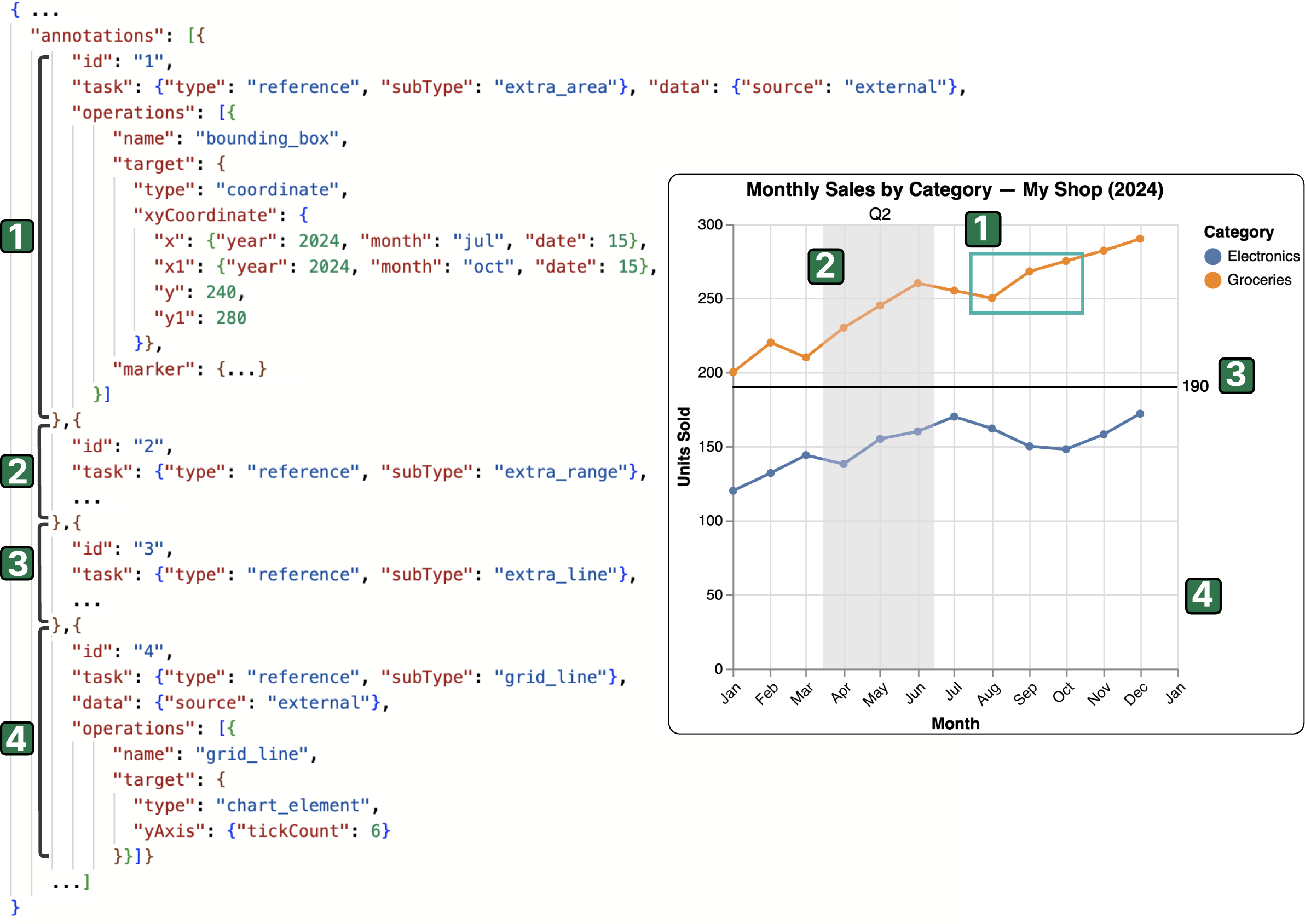}
    \vspace{ -1.5em}
    \caption{ 
ChartMark specification for a group line chart with four reference-type annotations: (\inlineicon{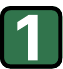}) a bounding box highlighting a specific time-value region (Jun-Oct 2024, 240-280 units), (\inlineicon{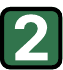}) a shaded background marking Q2, (\inlineicon{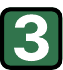}) a horizontal reference line at 190 units, and (\inlineicon{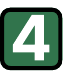}) customized y-axis grid lines with 6 intervals.
    } 
    \label{fig:Double_Line Example}
    \vspace{-1.5em}
\end{figure}

Using the representation framework as a foundation, we define the formal grammar of ChartMark. A unit ChartMark specification, denoted as \textit{annotatedChart}, consists of a base chart and one or more annotation elements:
\begin{equation}
\mathit{annotatedChart} := (\mathit{chart}, \mathit{annotations})
\end{equation}
\subsubsection{Chart Entity}  
The chart component represents the base chart and its properties:
\begin{equation}
\mathit{chart} := (\mathit{title}, \mathit{type}, \mathit{x\_data}, \mathit{y\_data}, \mathit{x\_name}, \mathit{y\_name},...)
\end{equation}

where $\mathit{type}$ indicates the chart type (e.g., bar, line, scatter), $*\_name$ parameters define the semantic labels for axes, and $*\_data$ parameters store the actual data values along each axis. 

\subsubsection{Annotation Elements}
Annotations are composed of a collection of annotation units, each with a specific purpose:
\begin{equation}
\mathit{annotations} := [\mathit{annotation}_1, \mathit{annotation}_2, \ldots, \mathit{annotation}_n]
\end{equation}
\begin{equation}
\mathit{annotation} := (\mathit{id}, \mathit{task}, \mathit{data}, \mathit{operations})
\end{equation}
where $\mathit{id}$ provides a unique identifier for each annotation, enabling referencing and composition of annotations.

\stitle{Task Component.} 
\revise{
The task component captures the purpose and intent of the annotation \cite{kong2012graphical, chen2022vizbelle}:
}
\begin{equation}
\mathit{task} := (\mathit{type}, \mathit{subType?})
\end{equation}
\begin{equation}
\mathit{type} := \mathit{reference} \; | \; \mathit{highlight} \; | \; \mathit{description} \; | \; \mathit{summary} \; | \; \mathit{trend} \; | \; \mathit{encoding}
\end{equation}

The optional $subType$ parameter is contingent on the primary task type and provides further specification. For example, when $type$ is ``summary'', $subType$ can be ``max'' or ``min'', which means the annotation is intended to highlight the location or elements representing maximum or minimum values in the chart.

\stitle{Data Component.}
\revise{
The data component specifies the information needed by the annotation \cite{rahman2024qualitative, hullman2013contextifier, bryan2016temporal}:
}
\begin{equation}
\mathit{data} := (\mathit{source}, \mathit{values})
\end{equation}
\begin{equation}
\mathit{source} := \mathit{external} \; | \; \mathit{derived} \; | \; \mathit{internal} \; | \; \mathit{none}
\end{equation}

The $\mathit{source}$ parameter indicates the origin of data, where $\mathit{external}$ represents information imported from outside the chart context, such as descriptive text and images (\inlineicon{figures/tag/orange/orange_3.pdf} in \cref{fig:Group_Scatter_Chart_Example}), as well as display configuration parameters such as tickcount settings in gridlines (\inlineicon{figures/tag/green/green_4.pdf} 
in \cref{fig:Double_Line Example}). $\mathit{derived}$ indicates computed values based on chart data, such asstatistical aggregates like mean (\cref{fig:teaser}-c). $\mathit{internal}$ refers to direct references to existing data items, used when annotations need to select data-related marks in the chart, such as in highlight annotations (\cref{fig:teaser}-a). $\mathit{none}$ is used for annotations that do not require specific data references.

Values represent a collection of multiple data item units, organized as an array. Each individual value unit is a tuple containing a type and either content or url:
\begin{equation}
\mathit{values} := [\mathit{value}_1, \mathit{value}_2, \ldots, \mathit{value}_n]
\end{equation}
\begin{equation}
\mathit{value} := (\mathit{type}, \mathit{content} \; | \; \mathit{url})
\end{equation}
\inlineicon{figures/tag/orange/orange_3.pdf} in \cref{fig:Group_Scatter_Chart_Example} shows two ways to specify data values: using a url to reference external images or using the content field to embed text directly. 

\revise{
\stitle{Operation Components.}
}
Operations are a collection of specific operating behaviors.
\revise{
Each operation represents a distinct annotation action that defines how visual elements are transformed \cite{chen2022vizbelle}:
}
\begin{equation}
\mathit{operations} := [\mathit{operation}_1, \mathit{operation}_2, \ldots, \mathit{operation}_n]
\end{equation}
\begin{equation}
\mathit{operation} := (\mathit{name}, \mathit{target}, \mathit{marker})
\end{equation}
\revise{
\stitle{Target Component.}
}
$target$ field specifies what elements the operation acts upon, 
\revise{which can be summarized as four types below \cite{ rahman2024qualitative, ren2017chartaccent, chen2022vizbelle}:
}
\begin{equation}
\mathit{target} := (\mathit{type}, \ldots)
\end{equation}
\begin{equation}
\mathit{type} := \mathit{data\_items} \; | \; \mathit{coordinate} \; | \; \mathit{chart\_element} \; | \; \mathit{annotation}
\end{equation}

Different target types require different additional parameters:

$data\_items$: uses filter parameters to select specific data points based on logical expressions (\inlineicon{figures/tag/orange/orange_3.pdf} in \cref{fig:Group_Scatter_Chart_Example}). 
$coordinate$: defines regions, intervals, points, or lines within the coordinate system using spatial parameters (\inlineicon{figures/tag/green/green_1.pdf} in \cref{fig:Double_Line Example}). 
$chart\_element$: references existing chart components for modification through component identifiers (\inlineicon{figures/tag/green/green_4.pdf} in \cref{fig:Double_Line Example}). 
$annotation$: references existing annotations through an ID parameter to enable composition of multiple annotations (\inlineicon{figures/tag/orange/orange_1.pdf} and \inlineicon{figures/tag/orange/orange_2.pdf} in \cref{fig:Group_Scatter_Chart_Example}).

\stitle{Marker Component.}
$marker$ field indicates that added or modified visual elements in a operation. It serves as a collection of display‐element options, represented as an optional parameter list, 
\revise{
where each parameter specifies a single, fine‐grained display‐element setting \cite{ren2017chartaccent}:
}
\begin{equation}
\mathit{marker} := (\mathit{line?}, \mathit{text?}, \ldots)
\end{equation}
\begin{equation}
\mathit{line} := (\mathit{size?}, \mathit{dashed?}, \ldots)
\end{equation}
\begin{equation}
    \mathit{text} := \ldots
\end{equation}
For example, when drawing a trend line, the detailed
display parameters like size or dashed for line can be set. (\inlineicon{figures/tag/orange/orange_1.pdf} in \cref{fig:Group_Scatter_Chart_Example})

\section{ChartMark Usage Demonstration}

To demonstrate the practical capabilities of the proposed ChartMark grammar, we developed a Python package named \texttt{ChartMark.py}, available as  
\revise{ 
open source at \href{https://github.com/HKUSTDial/ChartMark}{github.com/HKUSTDial/ChartMark}.
}
This toolkit facilitates converting ChartMark specifications into Vega-Lite visualizations~\cite{Satyanarayan2017} and provides users with an accessible interface to implement structured chart annotations seamlessly.

\subsection{Using ChartMark in Jupyter Notebook} 

\begin{figure}[t!]
	\centering	\includegraphics[width=\linewidth]{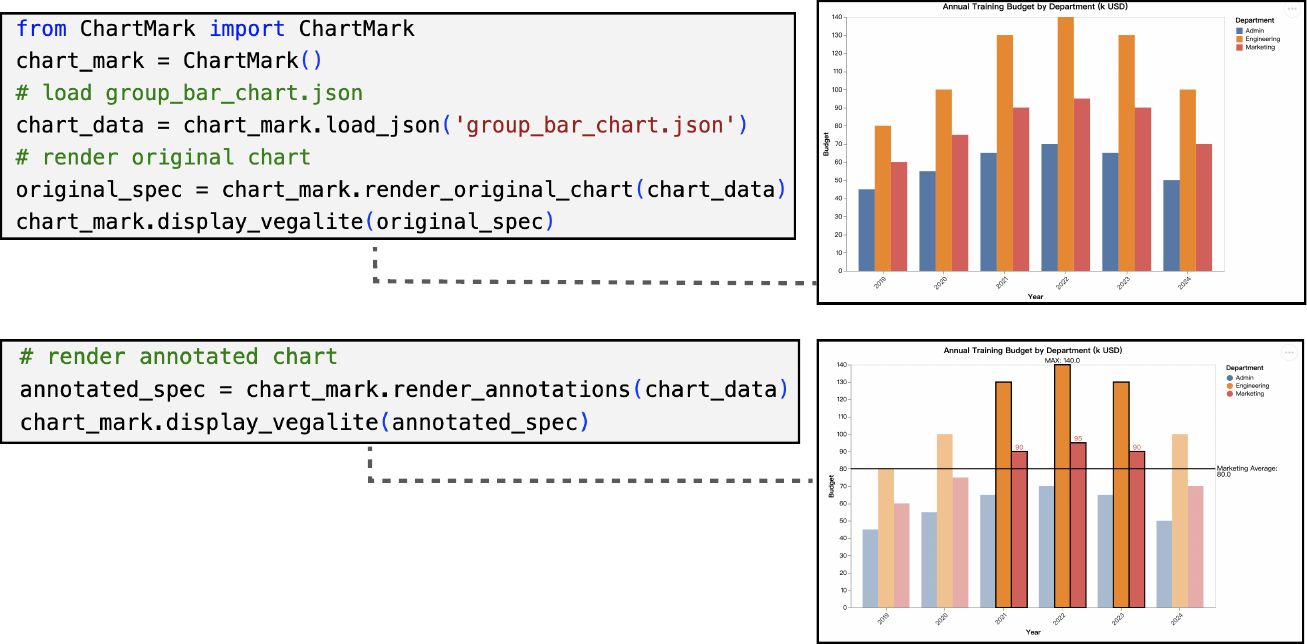}
    \vspace{-1.5em}
    \caption{ChartMark in Jupyter notebook} 
    \label{fig:code}
    \vspace{-1.5em}
\end{figure}

\cref{fig:code} illustrates a complete example of using \texttt{ChartMark.py} within a Jupyter notebook.
First, the users load a JSON-formatted ChartMark instance (see \cref{fig:teaser}) using \texttt{load\_json()}. Next, they generate a Vega-Lite specification for the base chart (without annotations) via \texttt{render\_original\_chart()}. To render this base chart in the notebook (upper half of~\cref{fig:code}), they pass the generated spec to \texttt{display\_vegalite()}. Then, the user applies annotations by calling \texttt{render\_annotations()}, which produces a fully annotated Vega-Lite specification based on the ChartMark grammar. Finally, they again use \texttt{display\_vegalite()} to display the annotated chart (lower half of \cref{fig:code}), showcasing various annotation types (e.g., highlighting, labeling, mean lines) that enhance the interpretability of the original chart. 

In addition, the package includes utility methods such as \texttt{get\_supported\_chart\_types()}, enabling users to quickly identify currently supported chart types.

\subsection{Extension and Reuse}
\revise{
Benefiting from a modular component-based architecture, ChartMark supports the rapid construction of custom annotations by reusing modular components. As an example, consider the creation of a custom annotation operation that adds stroke for specific slices in a pie chart (Annotation2 in \cref{fig:garmmar_overview}‑a). In \texttt{ChartMark.py}, creating such a custom annotation generally follows three steps:

\textbf{1. Decide the constructive components.} Select the necessary components that make up the new annotation node. While new Marker and Operation components need to be defined to represent the custom operation’s semantics and visual form, the Task, Data, and Target components can be reused directly due to their consistent roles with those in predefined annotation types.

\textbf{2. Register into the ChartMark AST.}
Add the node representing the new annotation type to the appropriate position in the abstract syntax tree (AST), and define parsing rules to interpret the content represented by this node.

\textbf{3. Implement transformation logic.}
Define how each operation within the annotation is mapped to concrete annotation statements in the target visualization language (e.g., Vega-Lite \cite{Satyanarayan2017}), ensuring correct rendering.
}

\section{Conclusion and Future Work}

We introduced ChartMark, a structured grammar designed to integrate fragmented annotation methods. ChartMark's hierarchical design can take into account both high-level annotation intent and low-level visual operations, and its definition is independent of specific visualization libraries. We verify its portability and powerful expressiveness based on Vega-Lite's implementation.
In future work, we plan to support dynamic annotation, explore new interactive methods that allow users to input in natural language and automatically generate annotations based on semantic intent~\cite{NLISurvey, Instructions}, and expand compatibility with more visualization libraries to enhance the accessibility of visualization and data storytelling capabilities~\cite{GEGraph,DVSurvey}.

\clearpage
\balance


\bibliographystyle{abbrv-doi}

\bibliography{main}

\begin{thebibliography}{10}

\bibitem{adobe_Illustrator}
Adobe illustrator. \url{https://www.adobe.com/products/illustrator}, 2025.

\bibitem{tableau}
Tableau. \url{https://www.tableau.com}, 2025.

\bibitem{amini2016authoring}
F.~Amini, N.~H. Riche, B.~Lee, A.~Monroy-Hernandez, and P.~Irani.
\newblock Authoring data-driven videos with dataclips.
\newblock {\em IEEE transactions on visualization and computer graphics}, 23(1):501--510, 2016.

\bibitem{bateman2010useful}
S.~Bateman, R.~L. Mandryk, C.~Gutwin, A.~Genest, D.~McDine, and C.~Brooks.
\newblock Useful junk? the effects of visual embellishment on comprehension and memorability of charts.
\newblock In {\em Proceedings of the SIGCHI conference on human factors in computing systems}, pp. 2573--2582, 2010.

\bibitem{Bostock2011}
M.~Bostock, V.~Ogievetsky, and J.~Heer.
\newblock {D3: Data-Driven Documents}.
\newblock {\em IEEE Transactions on Visualization and Computer Graphics}, 17(12):2301--2309, 2011.

\bibitem{Bostock2011D3}
M.~Bostock, V.~Ogievetsky, and J.~Heer.
\newblock D³ data-driven documents.
\newblock {\em IEEE Transactions on Visualization and Computer Graphics}, 17(12):2301--2309, 2011. doi: {{%
10\hspace{.1pt}\discretionary{.}{%
}{.}\hspace{.4pt}1109\discretionary{/}{%
}{/}TVCG\hspace{.1pt}\discretionary{.}{%
}{.}\hspace{.4pt}2011\hspace{.1pt}\discretionary{.}{%
}{.}\hspace{.4pt}185}}


\bibitem{brehmer2019timeline}
M.~Brehmer, B.~Lee, N.~H. Riche, D.~Tittsworth, K.~Lytvynets, D.~Edge, and C.~White.
\newblock Timeline storyteller.
\newblock In {\em Proceedings of the Computation+ Journalism Symposium, Miami, FL, USA}, vol.~6, 2019.

\bibitem{bryan2016temporal}
C.~Bryan, K.-L. Ma, and J.~Woodring.
\newblock Temporal summary images: An approach to narrative visualization via interactive annotation generation and placement.
\newblock {\em IEEE transactions on visualization and computer graphics}, 23(1):511--520, 2016.

\bibitem{chen2022vizbelle}
Q.~Chen, Z.~Liu, C.~Wang, X.~Lan, Y.~Chen, S.~Chen, and N.~Cao.
\newblock Vizbelle: A design space of embellishments for data visualization.
\newblock {\em arXiv preprint arXiv:2209.03642}, 2022.

\bibitem{eccles2008stories}
R.~Eccles, T.~Kapler, R.~Harper, and W.~Wright.
\newblock Stories in geotime.
\newblock {\em Information Visualization}, 7(1):3--17, 2008.

\bibitem{hao2024finflier}
J.~Hao, M.~Yang, Q.~Shi, Y.~Jiang, G.~Zhang, and W.~Zeng.
\newblock Finflier: Automating graphical overlays for financial visualizations with knowledge-grounding large language model.
\newblock {\em IEEE Transactions on Visualization and Computer Graphics}, pp. 1--17, 2024.

\bibitem{hullman2013contextifier}
J.~Hullman, N.~Diakopoulos, and E.~Adar.
\newblock Contextifier: automatic generation of annotated stock visualizations.
\newblock In {\em Proceedings of the SIGCHI Conference on human factors in computing systems}, pp. 2707--2716, 2013.

\bibitem{Hunter2007Matplotlib}
J.~D. Hunter.
\newblock Matplotlib: {A} 2d graphics environment.
\newblock {\em Computing in Science \& Engineering}, 9(3):90--95, 2007.

\bibitem{kong2012graphical}
N.~Kong and M.~Agrawala.
\newblock Graphical overlays: Using layered elements to aid chart reading.
\newblock {\em IEEE transactions on visualization and computer graphics}, 18(12):2631--2638, 2012.

\bibitem{Li2018ECharts}
C.~Li, J.~Zhao, B.~Lee, Q.~Yuan, and X.~Luo.
\newblock Echarts: A declarative framework for rapid construction of web-based visualization.
\newblock {\em Visual Informatics}, 2(1):1--12, 2018.

\bibitem{DBLP:conf/icde/LuoQ0018}
Y.~Luo, X.~Qin, N.~Tang, and G.~Li.
\newblock Deepeye: Towards automatic data visualization.
\newblock In {\em {ICDE}}, pp. 101--112. {IEEE} Computer Society, 2018.

\bibitem{moritz2018formalizing}
D.~Moritz, C.~Wang, G.~L. Nelson, H.~Lin, A.~M. Smith, B.~Howe, and J.~Heer.
\newblock Formalizing visualization design knowledge as constraints: Actionable and extensible models in draco.
\newblock {\em IEEE transactions on visualization and computer graphics}, 25(1):438--448, 2018.

\bibitem{rahman2024qualitative}
M.~D. Rahman, G.~J. Quadri, B.~Doppalapudi, D.~A. Szafir, and P.~Rosen.
\newblock A qualitative analysis of common practices in annotations: A taxonomy and design space.
\newblock {\em IEEE Transactions on Visualization and Computer Graphics}, pp. 360 -- 370, 2024.

\bibitem{ren2017chartaccent}
D.~Ren, M.~Brehmer, B.~Lee, T.~H{\"o}llerer, and E.~K. Choe.
\newblock Chartaccent: Annotation for data-driven storytelling.
\newblock In {\em 2017 IEEE Pacific Visualization Symposium (PacificVis)}, pp. 230--239. Ieee, 2017.

\bibitem{ren2018charticulator}
D.~Ren, B.~Lee, and M.~Brehmer.
\newblock Charticulator: Interactive construction of bespoke chart layouts.
\newblock {\em IEEE transactions on visualization and computer graphics}, 25(1):789--799, 2018.

\bibitem{Satyanarayan2017}
A.~Satyanarayan, D.~Moritz, K.~Wongsuphasawat, and J.~Heer.
\newblock {Vega-Lite: A Grammar of Interactive Graphics}.
\newblock {\em IEEE Transactions on Visualization and Computer Graphics}, 23(1):341--350, 2017.

\bibitem{Satyanarayan2014Declarative}
A.~Satyanarayan, K.~Wongsuphasawat, and J.~Heer.
\newblock Declarative interaction design for data visualization.
\newblock In {\em Proceedings of the 27th Annual ACM Symposium on User Interface Software and Technology (UIST)}, pp. 669--678. ACM, 2014. doi: {{%
10\hspace{.1pt}\discretionary{.}{%
}{.}\hspace{.4pt}1145\discretionary{/}{%
}{/}2642918\hspace{.1pt}\discretionary{.}{%
}{.}\hspace{.4pt}2647360}}


\bibitem{DVSurvey}
L.~Shen, H.~Li, Y.~Wang, and H.~Qu.
\newblock {Reflecting on Design Paradigms of Animated Data Video Tools}.
\newblock In {\em Proceedings of the 2025 CHI Conference on Human Factors in Computing Systems}, pp. 1--21. ACM, 2025.

\bibitem{Instructions}
L.~Shen, H.~Li, Y.~Wang, X.~Xie, and H.~Qu.
\newblock {Prompting Generative AI with Interaction-Augmented Instructions}.
\newblock In {\em Extended Abstracts of the CHI Conference on Human Factors in Computing Systems, CHI EA '25}, pp. 1--9. ACM, 2025.

\bibitem{NLISurvey}
L.~Shen, E.~Shen, Y.~Luo, X.~Yang, X.~Hu, X.~Zhang, Z.~Tai, and J.~Wang.
\newblock {Towards Natural Language Interfaces for Data Visualization: A Survey}.
\newblock {\em IEEE Transactions on Visualization and Computer Graphics}, 29(6):3121--3144, 2023.

\bibitem{taskvis}
L.~Shen, E.~Shen, Z.~Tai, Y.~Song, and J.~Wang.
\newblock {TaskVis: Task-oriented Visualization Recommendation}.
\newblock In {\em Proceedings of the 23th Eurographics Conference on Visualization, EuroVis'21}, pp. 91--95. Eurographics, 2021.

\bibitem{galvis}
L.~Shen, E.~Shen, Z.~Tai, Y.~Wang, Y.~Luo, and J.~Wang.
\newblock {GALVIS: Visualization Construction through Example-Powered Declarative Programming}.
\newblock In {\em Proceedings of the 31st ACM International Conference on Information {\&} Knowledge Management, CIKM'22}, pp. 4975--4979. ACM, 2022.

\bibitem{taskvisDSE}
L.~Shen, E.~Shen, Z.~Tai, Y.~Xu, J.~Dong, and J.~Wang.
\newblock {Visual Data Analysis with Task-Based Recommendations}.
\newblock {\em Data Science and Engineering}, 7(4):354--369, 2022.

\bibitem{GEGraph}
L.~Shen, Z.~Tai, E.~Shen, and J.~Wang.
\newblock {Graph Exploration With Embedding-Guided Layouts}.
\newblock {\em IEEE Transactions on Visualization and Computer Graphics}, 30(7):3693--3708, 2024.

\bibitem{shen2024ask}
S.~Shen, S.~Lu, L.~Shen, Z.~Sheng, N.~Tang, and Y.~Luo.
\newblock Ask humans or ai? exploring their roles in visualization troubleshooting.
\newblock {\em arXiv preprint arXiv:2412.07673}, 2024.

\bibitem{shi2025augmenting}
Y.~Shi, B.~Li, Y.~Luo, L.~Chen, and N.~Tang.
\newblock Augmenting realistic charts with virtual overlays.
\newblock In {\em Proceedings of the 2025 CHI Conference on Human Factors in Computing Systems}, pp. 1--23, 2025.

\bibitem{Siddiqui2016b}
T.~Siddiqui, A.~Kim, J.~Lee, K.~Karahalios, and A.~Parameswaran.
\newblock {Effortless data exploration with zenvisage: An expressive and interactive visual analytics system}.
\newblock {\em Proceedings of the VLDB Endowment}, 10(4):457--468, 2016.

\bibitem{Stolte2002b}
C.~Stolte, D.~Tang, and P.~Hanrahan.
\newblock {Polaris: a system for query, analysis, and visualization of multidimensional relational databases}.
\newblock {\em IEEE Transactions on Visualization and Computer Graphics}, 8(1):52--65, 2002.

\bibitem{Wickham2010}
H.~Wickham.
\newblock {A Layered Grammar of Graphics}.
\newblock {\em Journal of Computational and Graphical Statistics}, 19(1):3--28, 2010.

\bibitem{Wickham2016ggplot2}
H.~Wickham.
\newblock {\em ggplot2: Elegant Graphics for Data Analysis}.
\newblock Springer International Publishing, 2 ed., 2016.

\bibitem{wilkinson2012grammar}
L.~Wilkinson.
\newblock The grammar of graphics, in ‘handbook of computational statistics’, 2012.

\bibitem{wu2025chartcards}
Y.~Wu, L.~Yan, L.~Shen, Y.~Mei, J.~Wang, and Y.~Luo.
\newblock Chartcards: A chart-metadata generation framework for multi-task chart understanding.
\newblock {\em arXiv preprint arXiv:2505.15046}, 2025.

\bibitem{wu2024chartinsights}
Y.~Wu, L.~Yan, L.~Shen, Y.~Wang, N.~Tang, and Y.~Luo.
\newblock {ChartInsights: Evaluating Multimodal Large Language Models for Low-Level Chart Question Answering}.
\newblock In {\em Findings of the Association for Computational Linguistics: EMNLP 2024}, pp. 12174--12200. ACL, Stroudsburg, PA, USA, 2024.

\bibitem{wu2025boosting}
Y.~Wu, L.~Yan, Y.~Zhu, Y.~Mei, J.~Wang, N.~Tang, and Y.~Luo.
\newblock Boosting text-to-chart retrieval through training with synthesized semantic insights.
\newblock {\em arXiv preprint arXiv:2505.10043}, 2025.

\bibitem{xie2024haichart}
Y.~Xie, Y.~Luo, G.~Li, and N.~Tang.
\newblock Haichart: Human and ai paired visualization system.
\newblock {\em Proceedings of the VLDB Endowment}, 17(11):3178--3191, 2024.

\bibitem{yang2024askchart}
X.~Yang, Y.~Wu, Y.~Zhu, N.~Tang, and Y.~Luo.
\newblock Askchart: Universal chart understanding through textual enhancement.
\newblock {\em arXiv preprint arXiv:2412.19146}, 2024.

\bibitem{PyGWalker}
Y.~Yu, L.~Shen, F.~Long, H.~Qu, and H.~Chen.
\newblock {PyGWalker: On-the-fly Assistant for Exploratory Visual Data Analysis}.
\newblock In {\em Proceedings of IEEE Visualization and Visual Analytics, IEEE VIS'24}, pp. 1--5, 2024.

\end{thebibliography}
\end{document}